# How should a fixed budget of dwell time be spent in scanning electron microscopy to optimize image quality?

Authors: Patrick Trampert[1,2], Faysal Bourghorbel[3], Pavel Potocek[3], Maurice Peemen[3], Christian Schlinkmann[1,2], Tim Dahmen[1,*] and Philipp Slusallek[1,2]

[1] *German Research Center for Artificial Intelligence GmbH (DFKI), 66123 Saarbrücken, Germany*
[2] *Saarland University, 66123 Saarbrücken, Germany*
[3] *FEI, Eindhoven, Netherlands*
[*]) *email: Tim.Dahmen@dfki.de*

**Abstract**

In scanning electron microscopy, the achievable image quality is often limited by a maximum feasible acquisition time per dataset. Particularly with regard to three-dimensional or large field-of-view imaging, a compromise must be found between a high amount of shot noise, which leads to a low signal-to-noise ratio, and excessive acquisition times. Assuming a fixed acquisition time per frame, we compared three different strategies for algorithm-assisted image acquisition in scanning electron microscopy. We evaluated (1) raster scanning with a reduced dwell time per pixel followed by a state-of-the-art Denoising algorithm, (2) raster scanning with a decreased resolution in conjunction with a state-of-the-art Super Resolution algorithm, and (3) a sparse scanning approach where a fixed percentage of pixels is visited by the beam in combination with state-of-the-art inpainting algorithms. Additionally, we considered increased beam currents for each of the strategies. The experiments showed that sparse scanning using an appropriate reconstruction technique was superior to the other strategies. Adapting the reconstruction to a given situation to optimize quality played a major role.

## Introduction

Scanning electron microscopes (SEM) play a central role for analyzing the composition of materials and structures by imaging centimeters of a sample at nanometer scales. At such a fine resolution an analysis of samples requires a vast amount of data, which can take months of recording [1]. A conventional recording scheme for SEM is the raster scan, i.e. scanning on a regular grid, where each pixel location of a sample is visited for a given amount of dwell time per pixel that corresponds to a desired signal-to-noise ratio (SNR). For instance, the pure acquisition time via raster scan of a cube with edge length 0.02 mm and a resolution of 5 *nm* per pixel, which results in a volume resolution of $4096^3$, takes nearly eight days when the dwell time per pixel is 10 *μs*. This example shows that the data acquisition of cubes with an edge length of a millimeter with the same parameters is infeasible as raster scan.

**Strategy 1: Reduced Dwell Time Raster Scan plus Denoising**
There are different approaches that can reduce the acquisition time of a volume. The simplest way of reducing the dwell time per pixel scales linearly with the whole acquisition time. However, as each pixel value is determined from a decreasing number of electrons, the statistics of the measurements leads to decreased precision manifested as shot noise. This problem can be addressed in software by applying a Denoising algorithm to the image. Recently, algorithms that use prior knowledge on the form of previously learned dictionaries have been established as state-of-the art in Denoising. One such algorithm, inspired by Compressive Sensing (CS) is Geometric analysis operator learning (GOAL) [2]. In GOAL, image reconstruction underlies the assumption that natural images have a sparse representation over a so called dictionary. A dictionary is a collection of small image patches of a fixed size, e.g. 8×8 pixels. GOAL uses the dictionary as basis of the image patch space, so that linear combinations of only few dictionary entries resemble patches to be inserted. Instead of using mathematical models like Wavelets or Curvlets as dictionary, GOAL uses a representative training set to learn a maximally sparse representation over this training set as dictionary. This is done via convex optimization and conjugate gradients. The learned dictionary, respectively analysis operator, is then used to denoise an image.

**Strategy 2: Low Resolution Raster Scan plus Super Resolution**
Rather than reducing the dwell time per pixel, it is also possible to decrease the number of pixels. By increasing the pixel size and scanning the specimen on a regular grid in larger steps, the sample is discretized to fewer pixel values and acquisition time can be substantially lowered while maintaining the same measurement precision per pixel. Obviously, image resolution is lost as high frequency details in the image fall below the Nyquist threshold. This issue can be addressed by Super Resolution (SR) approaches, which may help to retrieve the original resolution. Given a low resolution image of size M×N, SR image reconstruction approaches aim to increase the resolution to a higher resolution image of, for example, size 2M×2N, while not losing quality. In our study, the method from Villena et al. [3], which is based on the robust SR method using bilateral Total Variation priors [4], was applied. The authors used a hierarchical Bayesian framework to minimize a linear convex combination of the Kullback-Leibler divergence to find a unique approximation of the high resolution image.

**Strategy 3: Random Sparse Scan plus Inpainting**
Instead of reducing dwell time per pixel or increasing pixel size, it is also possible to reduce the number of pixels that are scanned without increasing the pixel size. This is done by randomly selecting a subset of the image pixels and scanning only this subset. These pixels are then used to reconstruct the image using inpainting algorithms. A larger number of inpainting methods has been proposed including relatively straight forward interpolation methods, methods inspired by Compressive Sensing theory and Exemplar Based Inpainting methods inspired by image editing. We evaluated several inpainting methods: (1) Natural neighbor interpolation [5], which is a simple interpolation method for scattered data based on Voronoi tessellation, (2) Beta Process Factor Analysis (BPFA) and (3) GOAL, which both are inpainting algorithms inspired by CS, and (4) Exemplar Based Inpainting.

**Strategy 3.1: Natural Neighbor Interpolation**
Given a sparsely sampled image, unknown pixels are inserted by weighting the areas of surrounding Voronoi cells. This is done by inserting one Voronoi cell at a time for a new pixel, whose value is a weighted average of the Voronoi cells originally present in the area of the new cell. This kind of interpolation delivers smother results than bilinear or bicubic interpolation.

**Strategies 3.2 and 3.3: Compressive Sensing Inpainting**
Beneath interpolation, there are more sophisticated methods for inpainting that are inspired by ideas from CS [6]. Binev et al. [7] have examined, which parts of CS may be useful for EM. The authors analyzed different setups in EM where CS may have a useful impact. Stevens et al. [8] applied CS via Bayesian dictionary learning to high-resolution scanning transmission electron microscopy (STEM) images. The BPFA algorithm is a nonparametric Bayesian method that has been extended for image processing applications by Zhou et al. [9]. Given incomplete measurements, spatial inter-relationships are exploited using the Dirichlet process and probit-stick-breaking priors. This means that known pixels of the sparse image are used to infer an appropriate dictionary that is used for reconstruction of the missing pixels. It was shown that 5% of the pixels of an image can suffice to recover the original image by applying inpainting. Anderson et al. [10] have demonstrated sparse sampling in an operational SEM. They were able to speed up acquisition time by a factor of three while preserving an acceptable image quality. A randomly selected subset of pixels was used to reconstruct the original image by applying a split-Bregman formulation of regularized basis pursuit that leveraged block discrete cosine transform as a sparsifying basis, which is a CS method. GOAL [2] can also be used to recover sparsely sampled data via inpainting.

**Strategy 3.4: Exemplar Based Inpainting**
A further approach for the reconstruction of sparse data is Exemplar Based Inpainting [11], which is a class of inpainting algorithms known from image processing. There, the method is used to restore damaged regions of images or to remove objects from images by inserting information from the surroundings. Missing values are modified, so that the inserted parts match the surroundings visually. Adapting this idea, we developed a three dimensional inpainting method [12] that inserts voxels with the help of prior knowledge. Instead of using the surroundings of missing voxels, a dictionary is trained based on uncorrupted data. This dictionary contains a large number of image patches with a predefined size from comparable images to the microscope data that has to be reconstructed. Missing voxels are reconstructed iteratively by inserting whole patches. The procedure starts with identifying a region in the sparse image where a patch should be inserted. The scanned voxels in that patch are then used to find a patch in the dictionary that fits best. A cost function is used to determine this patch, for example the $L_2$ norm of known voxels. Missing voxels are inserted into the sparse image until the whole image has been reconstructed.

**Orthogonal Strategy: Beam Current**

Increasing the number of electrons per pixel without increasing the dwell time is an additional option, which can only be achieved by an increased beam current. This has the advantage of improving electron statistics without inducing additional acquisition time. However, the disadvantage is that the virtual spot size of the electron gun increases and the image appears blurred. The influence of increased beam current is evaluated in combination with each of the presented strategies.

To summarize, we assume a fixed budget of total dwell time for the acquisition of an image. We compare three strategies to acquire the best possible images with the given budget. Furthermore, we evaluate each of the strategies with different beam currents. The strategies are: (1) scan each pixel in a raster scan with shorter dwell time and address the resulting noise using Denoising algorithms, (2) increase the pixel size and pixel spacing and address the reduced resolution using Super Resolution algorithms, and (3) scan a randomly selected subset of all pixels and synthetically generate the missing data using inpainting algorithms. Instead of concentrating on low dose imaging[13], we fixed the available time per frame for the whole acquisition in this study. We ask the question how this budget should be distributed to maximize the information content of final reconstructions.

## Materials and Methods

### Datasets used for evaluation

We acquired nano-scale cell biology images on a scanning electron microscope (SEM), the FEI Helios 650 dual beam system. We used the optical ultra high resolution (UHR) mode at 4 *mm* working distance in high vacuum. The electron landing energy was 2 *kV* using the through the lens detector (TLD) in the back scatter electron (BSE) detection mode. A block specimen of the mouse brain was selected for imaging. The same frame was scanned repeatedly at a resolution of 1024×884 pixels using different dwell times of 10 *μs* and 30 *μs* and different beam currents of 0.1 *nA*, 0.2 *nA*, 0.4 *nA*, and 0.8 *nA*. After image registration we selected varying regions of interest within these scans with a size of 480×424 for further processing.

Based on those datasets, we simulated all acquisition schemes investigated in this study as follows. Combining the 10 *μs* and 30 *μs* images for each spot, we synthetically generated 40 *μs* images. One spot with 40 *μs* dwell time and 0.1 *nA* beam current was chosen as ground truth for the evaluation. This ground truth was also slightly smoothed with Gaussian smoothing using $\sigma = 0.5$. This was done as it is quite common to post-process microscope acquisitions before using the images for further processing to remove small perturbations caused by noise. Low resolution scans were simulated by downscaling the images by a factor of two. Random sparse scans were simulated by blanking all but a randomly selected set of pixels and setting the corresponding pixel values to not-a-number. The different strategies and corresponding reconstruction methods were applied to images of the selected spot using different beam currents, as mentioned before, while keeping the average dwell time per pixel for each acquisition constant at 10 *μs*. This corresponds to an acquisition time of around two seconds per spot. The datasets are provided as supplementary S1.

**Algorithmic Setups**

We investigated different approaches to acquire data from a SEM given a fixed acquisition time per frame. (1) Raster scans with a reduced dwell time per pixel of 10 *μs* (Figure 1b) and a pixel size of 5 *nm* were acquired. The unprocessed images are called "Original Raster", a version processed using GOAL for Denoising is denoted "GOAL Denoising". (2) Raster scans with half of the original resolution were simulated from the corresponding full frame images at 40 *μs* dwell time and 5 *nm* pixel size (Figure 1c), which corresponds to 10 *nm* pixel size. The resulting images were then upscaled using the described SR approach from [3] to increase the resolution to the original size. The corresponding dataset is denoted "Super Resolution". (3) Sparse scans with 25% randomly selected pixels were simulated from the 40 *μs* dwell time images at 5*nm* pixel size (Figure 1d). The resulting images were reconstructed using different inpainting methods, which includes interpolation. Linear, bicubic, nearest neighbor, and natural neighbor interpolation were applied. As natural neighbor interpolation proved to be the best interpolation method for the data used in this study, only natural neighbor interpolation results are shown (NN Interpolation). The compared inpainting methods were Exemplar Based Inpainting (EBI), GOAL, and BPFA.

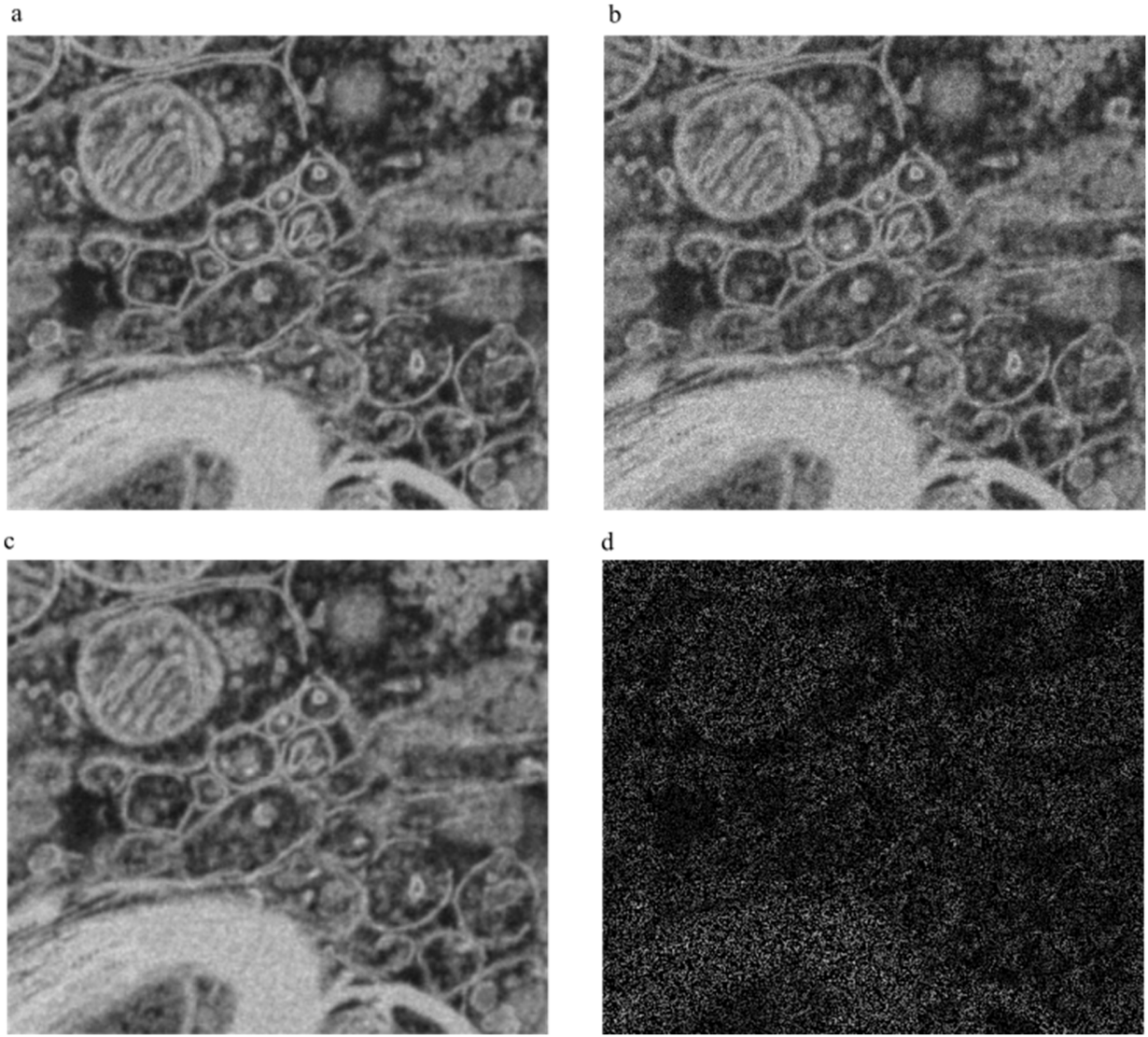

*Figure 1. a) Ground truth at 40 µs. b) Raster scan with a reduced dwell time per pixel of 10 µs and a pixel size of 5 nm. c) Raster scan with a dwell time per pixel of 40 µs and an increased pixel size of 10 nm (downscaled image containing only 25% pixels from a ). d) Sparse scan with a dwell time per pixel of 40 µs and a pixel size of 5 nm that only contains 25% of randomly selected pixels.*

**Evaluation Method**

We found that commonly used evaluation methods such as Peak Signal to Noise Ratio (PSNR) and Structural Similarity Index (SSIM) [16] did not correspond to the visual quality of the final images in a reliable way. Therefore, we additionally used the Complex Wavelet Structural Similarity (CW-SSIM) [14] that is insensitive to geometric distortions that do not influence structure, which is an important property for the application in this study. For similar reasons, we included PSNR-HVS-M [15], which is an adapted Peak Signal to Noise Ratio [16] that uses a between-coefficient contrast masking of discrete cosine transform basis functions. This enhances the capabilities of the metric to capture the visual appearance to the human visual system, which has been shown in [17] by conducting subjective experiments. As demonstrated by Wang and Bovik [18], quantitative measures may be very misleading. Therefore, the visual comparisons may be more important in this study than the quantitative measures.

## Results

**Reconstruction Quality**

In order to investigate how well the different acquisition methods preserve small structures, we manually selected 30 different locations in the dataset. A quantitative evaluation of the reconstruction quality is provided in Table 1. Reconstructed data for all structures is provided as supplementary S2.

*Table 1. Quantitative performance metrics applied to datasets obtained using different sampling strategies and reconstruction algorithms. 30 structures, like the ones in Figure 2, were evaluated and results are given as mean ± standard deviation σ. For each metric, the method with the highest value and all methods within ±1σ are indicated in bold letters.*

| **Method** | **PSNR** | **PSNR-HVS-M** | **SSIM** | **CW-SSIM** |
|---|---|---|---|---|
| Original Raster | 25.57 ± 0.60 | **27.35 ± 0.96** | 0.726 ± 0.061 | **0.890 ± 0.056** |
| GOAL Denoising | 27.20 ± 0.91 | 23.81 ± 1.06 | **0.888 ± 0.038** | **0.905 ± 0.050** |
| Super Resolution | 25.48 ± 0.71 | 21.37 ± 0.91 | 0.814 ± 0.047 | **0.886 ± 0.056** |
| NN Interpolation | 27.38 ± 1.18 | 23.91 ± 1.36 | **0.885 ± 0.034** | **0.891 ± 0.046** |
| GOAL Inpainting | 27.37 ± 1.08 | 23.85 ± 1.19 | **0.894 ± 0.034** | **0.894 ± 0.050** |
| EBI | 25.86 ± 0.97 | 23.01 ± 1.24 | 0.789 ± 0.053 | 0.829 ± 0.064 |
| BPFA Inpainting | **31.34 ± 0.77** | **29.22 ± 1.42** | **0.885 ± 0.040** | **0.891 ± 0.044** |

A conventional raster scan with a dwell time of 10 μs per pixel was used as a baseline for the results. This approach resulted in a PSNR of 25.57 ± 0.60, PSNR-HVS-M of 27.35 ± 0.96, SSIM of 0.726 ± 0.061, and CW-SSIM of 0.890 ± 0.056. Applying a state-of-the art Denoising using the GOAL operator improved only some of the metrics (PSNR = 27.20 ± 0.91) but reduced PSNR-HVS-M to 23.81 ± 1.06.

Processing a low resolution image of 10 *nm* pixel size with a state-of-the art SR algorithm gave inconclusive results as well. PSNR, SSIM and CW-SSIM were statistically the same, PSNR-HVS-M was worse compared to the original raster scan.

Better results were obtained by randomly selecting 25% of the pixels at 5 *nm* pixel size and 40 *μs* pixel dwell time, followed by a reconstruction using inpainting algorithms. While nearest neighbor interpolation, GOAL Inpainting and EBI showed slightly better performance than the original raster scan, all three methods were statistically indistinguishable from the smoothed version obtained by GOAL Denoising. Best results were obtained using BPFA Inpainting. The method resulted in the highest PSNR value of 31.34 ± 0.77, the highest PSNR-HVS-M of 29.22 ± 1.42 and was among the highest SSIM and CW-SSIM values.

We found that the different measures did not agree in all cases, nor did the quantitative measures always correspond to our subjective impression of image quality. Therefore, results for some representative structures are shown in Figure 2. As can be seen, the applied acquisition and reconstruction strategies generate visually different results.

A particularly challenging structure is shown in Figure 2, row 6. The image excerpt shows a closed, droplet-like structure with a clear interior. The results from GOAL Denoising and Super Resolution show a deviating structure, where the interior is not identifiable, nor is the structure closed. In the results obtained with EBI, a closed structure is visible. However, the interior is not clearly identifiable as it could also be noise and it is smaller than in the ground truth. In the results obtained with BPFA Inpainting, the structure is reconstructed more faithfully in terms of a closed structure and a clear interior.

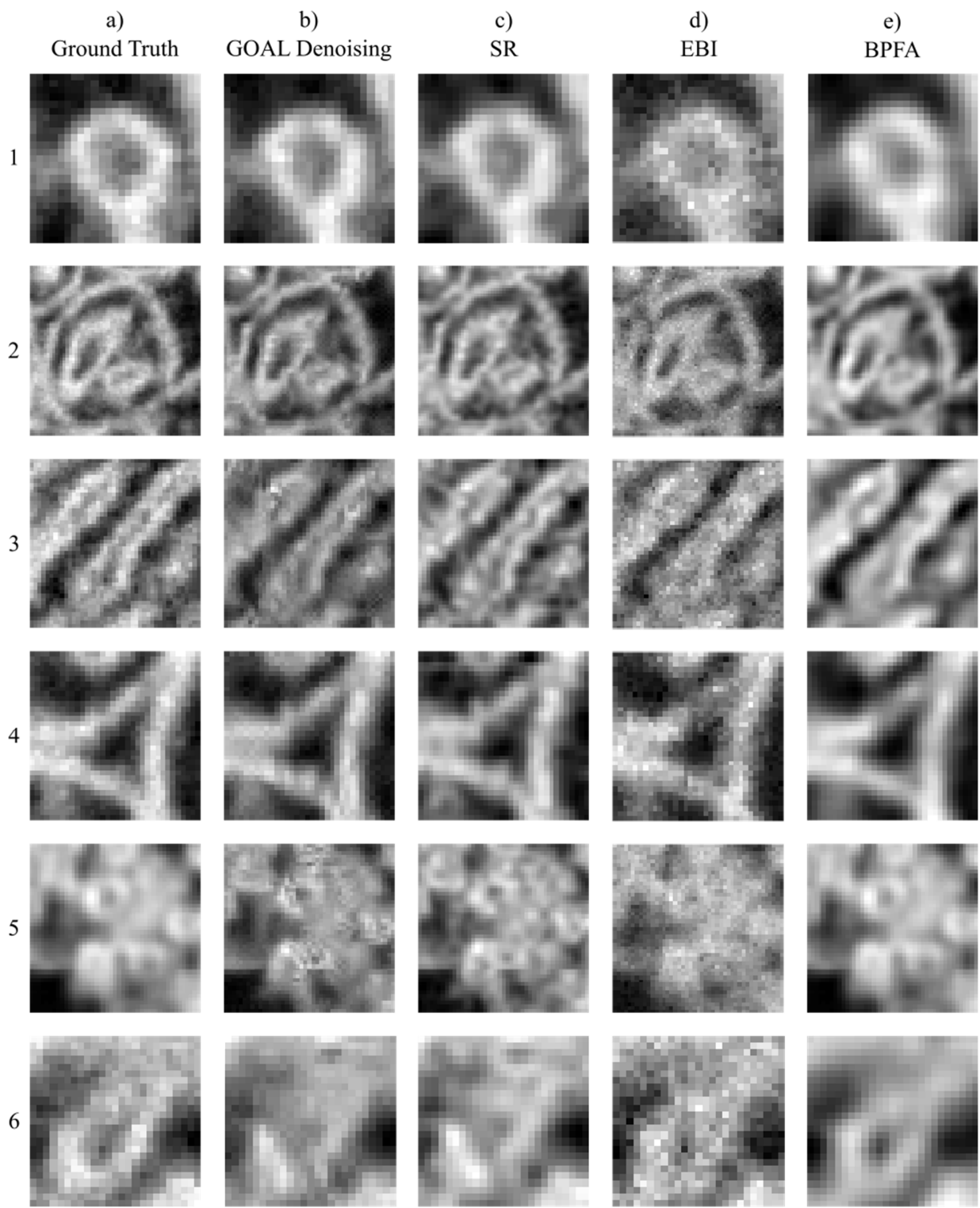

*Figure 2. Impact of the different acquisition strategies and reconstruction algorithms shown on some representative structures. a) Ground truth (raster scan at 5 nm pixel size and 40 µs dwell time per pixel). b) Raster scan acquired at 10µs dwell time per pixel, denoised using GOAL. c)*

*Low resolution image acquired at 10 nm pixel size and 40 µs dwell time per pixel, processed using a Super Resolution algorithm. d) Random sparse scan of 25% pixels at 5 nm pixel size and 40 µs dwell time per pixel, reconstructed using exemplar based inpainting. e) Acquisition as in d), but reconstructed using BPFA. Note that columns b)-e) use the same total per frame acquisition time budget of 10 µs dwell time per pixel on average.*

**Influence of Beam Current**

In addition to investigating the individual acquisition approaches and corresponding reconstruction methods in terms of dwell time, different beam currents were evaluated (Table 2). We incrementally increased the beam current from 0.1 *nA* to 0.2 *nA*, 0.4 *nA*, and 0.8 *nA* and repeated the previously described reconstruction experiment for each setting. Results in terms of PSNR-HVS-M are shown in Table 2, additional measures are provided as supplementary S3.

Optimal results were obtained for 0.1 *nA* for all reconstruction methods. Increasing the beam current either deteriorated results or gave the same quality within measurement precision.

*Table 2. Influence of the beam current on the reconstruction quality, measured in terms of PSNR-HVS-M values. 30 structures, like the ones in Figure 2, were evaluated and results are given as mean ± standard deviation.*

| Algorithm | 0.1nA | 0.2nA | 0.4nA | 0.8nA |
|---|---|---|---|---|
| Original Raster | **27.35 ± 0.96** | 24.41 ± 1.02 | 22.10 ± 1.38 | 19.48 ± 1.46 |
| GOAL Denoising | **23.81 ± 1.06** | **22.82 ± 1.86** | **21.18 ± 1.72** | 19.08 ± 1.54 |
| Super Resolution | **21.37 ± 0.91** | **22.05 ± 1.59** | **21.16 ± 1.56** | **19.02 ± 1.46** |
| NN Interpolation | **23.91 ± 1.36** | **22.14 ± 1.73** | 20.57 ± 1.75 | 18.67 ± 1.49 |
| GOAL Inpainting | **23.85 ± 1.19** | **22.60 ± 1.77** | 21.03 ± 1.85 | 18.92 ± 1.53 |
| EBI | **23.01 ± 1.24** | **21.33 ± 1.53** | 19.95 ± 1.54 | 18.30 ± 1.44 |
| BPFA Inpainting | **29.22 ± 1.42** | 24.36 ± 1.26 | 21.80 ± 1.54 | 19.26 ± 1.57 |

**Influence of Inpainting Dictionary**

Exemplar Based Inpainting is very sensitive to the dictionary used for the reconstruction. We therefore tested different dictionaries ranging from a very small dictionary as worst case scenario to an theoretical ideal dictionary built from ground truth data. This evaluation showed that an increasing dictionary quality lead to higher quality reconstructions. In the course of this study we achieved the following measure values as upper bounds on the impact of the dictionary on the reconstruction quality: PSNR = 38.89 ± 1.55, PSNR-HVS-M = 49.40 ± 5.57, SSIM = 0.9823 ± 0.0061, and CW-SSIM = 0.9976 ± 0.022. While an investigation into choosing an optimal dictionary is an interesting question, we will not go into further detail here, as the reconstruction method itself is out of scope for this study. As opposed to exemplar based inpainting, GOAL is quite insensitive to the used dictionary. During tests of dictionaries learned on different datasets we found that for SEM reconstructions there were only small differences in terms of visual quality, which means that GOAL Inpainting results did not depend much on the data used for dictionary learning. For BPFA the dictionary is always optimal, because this method learns the used dictionary directly from the sparse data it is applied to for reconstruction, so there is only one result for BPFA Inpainting.

## Discussion

We have collected SEM data applying different sampling and reconstruction strategies with a fixed budget of acquisition time per frame. The compared strategies were: (1) A raster scan at reduced dwell time per pixel, followed by image processing using a state-of-the-art Denoising algorithm. (2) A raster scan at increased pixel size followed by resolution enhancement using a state-of-the-art Super Resolution algorithm. (3) Sparse image acquisition where only a randomly selected subset of pixels is acquired, followed by a reconstruction using a variety of inpainting algorithms, which also includes interpolation methods. As an orthogonal technique, we investigated the result of increasing the beam current.

Looking at the quantitative measures and at the image structures (Figure 1), it became evident that increasing the beam current for quality enhancement was counter-productive for all approaches investigated in this study. The reason for this is that at higher beam currents the virtual spot size increases, which leads to a defocused beam and thus to smoother and blurrier acquisitions. We found that in our setup, this blurring is not compensated for by the reduced shot noise. Furthermore, as a higher beam current also means a higher electron dose, increasing the beam current may damage the biological sample, which in turn may lead to distortions in the acquisition. Consequently, we excluded the higher beam currents from further evaluations and only took the 0.1 nA beam current results for further assessment. However, while the increase of the virtual spot size at higher beam currents is a fundamental principle, the strength of the effect depends on technical parameters such as the construction of the electron gun, the column, and the acceleration voltage. Therefore, the optimal dwell time that balances beam blurring against shot noise might differ among microscope setups.

Concerning the optimal acquisition strategy, we found that a raster scan at reduced pixel dwell time can be improved by means of GOAL Denoising. However, the effect was less pronounced than we expected and did not show consistently across all quantitative measures for image quality.

Comparable results were obtained by using a raster scan at increased pixel size, followed by the application of a Super Resolution algorithm. In general, we found that quantitative ways to determine image quality can vary between measures. As visually different reconstructions can have almost the same values in many cases, one must be careful when interpreting small differences in these numbers. Keeping this in mind, we consider most results for random sparse scanning followed by inpainting to be inconclusive as well. The reconstructions using NN interpolation, Exemplar Based Inpainting and GOAL Inpainting show similar results as the GOAL Denoising method in the sense that they are superior to the original raster scan in some error metrics but not in others, and seem visually superior or not depending on what structure is inspected.

A clearly different result is obtained for a random sparse scan followed by a reconstruction using BPFA Inpainting, which gives the best results among the compared methods. This finding is supported by both the evaluation measures and the visual inspection of the structures.

One important consideration for dictionary based inpainting algorithms such as BPFA, GOAL and EBI is the impact the dictionary has on the reconstruction quality. BPFA performs on-the-fly learning of a dictionary from the sparse dataset, and thus is independent of the provided dictionary. GOAL and EBI, on the other hand, rely on additional prior knowledge that is provided in the form of a previously acquired dictionary. We found that GOAL performs rather stable when using different dictionaries while EBI is highly sensitive to the used prior for dictionary learning. The optimal structure for EBI dictionaries for inpainting sparse scanning data is not understood well today, so there may be a large potential for improvement in this direction.

## Conclusion

We have evaluated different strategies to acquire SEM data when a fixed budget of acquisition time per frame is used. Raster scanning with reduced dwell time per pixel can be improved by state-of-the art Denoising algorithms. Raster scanning with increased pixel size followed by a state-of-the art Super Resolution algorithm did not increase image quality compared to a conventional raster scan. Sparse datasets, where only a randomly selected set of pixels is scanned, can be reconstructed using inpainting methods such as interpolation, exemplar based inpainting or inpainting methods inspired by Compressive Sensing. While all investigated inpainting methods resulted in superior results compared to a raster scan, best results were obtained using BPFA. We conclude that the conventional approach of image acquisition should be challenged. Sparse acquisition techniques should have a much larger role in future microscopy, particularly for situations such as very large field-of-view scanning and three-dimensional acquisitions of biological samples, where the total acquisition time and electron dose are limiting factors.

## Acknowledgement

The authors thank FEI for funding the research that lead to this study and for providing the necessary infrastructure and the DFKI GmbH for additional funding.